\documentclass[10pt,a4paper,openbib]{article}
\usepackage[latin1]{inputenc}
\usepackage{amsmath, amsthm}
\usepackage{amsfonts}
\usepackage{amssymb}
\usepackage{graphicx}
\usepackage[colorlinks=false,
            linkcolor=blue,
            anchorcolor=blue,
            citecolor=green,
            ]{hyperref}
\usepackage{xcolor}
\usepackage{algorithm}
\usepackage{algorithmic}
\usepackage{cite}
\usepackage{indentfirst}
\usepackage{etoolbox}
\begin{document}
\title{Comprehend DeepWalk as Matrix Factorization}
\author{Cheng Yang\\
Tsinghua University\\
cheng-ya14@mails.tsinghua.edu.cn\\
\and
Zhiyuan Liu\\
Tsinghua University\\
liuzy@tsinghua.edu.cn\\
}
\maketitle
\begin{abstract}
Word2vec\cite{mikolov2013efficient}, as an efficient tool for learning vector representation of words has shown its effectiveness in many natural language processing tasks. Mikolov et al. issued Skip-Gram and Negative Sampling model\cite{mikolov2013distributed} for developing this toolbox. Perozzi et al. introduced the Skip-Gram model into the study of social network for the first time, and designed an algorithm named DeepWalk\cite{perozzi2014deepwalk} for learning node embedding on a graph. We prove that the DeepWalk algorithm is actually factoring a matrix $M$ where each entry $M_{ij}$ is logarithm of the average probability that node $i$ randomly walks to node $j$ in fix steps. We will explain it in section $3$.
\end{abstract}

\section{Notation}
Network $G=(V,E)$. Node-context set $D$ is generated from random walk, where each piece of $D$ is a node-context pair $(v,c)$. $V$ is the set of nodes and $V_C$ is the set of context nodes. In most cases, $V=V_C$.

Consider a node-context pair $(v,c)$:

$\#(v,c)$ denotes the number of times $(v,c)$ appears in $D$. $\#(v)=\sum_{c'\in V_C}\#(v,c')$ and $\#(c)=\sum_{v'\in V}\#(v',c)$ denotes the number of times $v$ and $c$ appears in $D$. Note that $|D|=\sum_{v'\in V}\sum_{c'\in V_C}\#(v',c')$.

DeepWalk algorithm embeds a node $v$ into a $d$-dimension vector $\overrightarrow{v}\in \mathbb{R}^d$. Also, a context node $c\in V_C$ is represented by a $d$-dimension vector $\overrightarrow{c}\in \mathbb{R}^d$. Let $W$ be a $|V|\times d$ matrix where row $i$ is vector $\overrightarrow{v_i}$ and $H$ be a $|V_C|\times d$ matrix where row $j$ is vector $\overrightarrow{c_j}$. Our goal is to figure out a matrix $M=WH^T$.
\section{Proof}
Perozzi et al. implemented DeepWalk algorithm with Skip-Gram and Hierarchical Softmax model. Note that Hierarchical Softmax\cite{morin2005hierarchical}\cite{mnih2009scalable} is a variant of softmax for speeding the training time. In this section, we give proofs for both Negative Sampling and softmax with Skip-Gram model.
\\

\subsection{Negative Sampling}
Negative Sampling approximately maximizes the probability of softmax function by randomly choosing $k$ negative samples from context set. Levy and Goldberg showed that Skip-Gram with Negative Sampling model(SGNS) is implicitly factorizing a word-context matrix\cite{levy2014neural} by assuming that dimensionality $d$ is sufficiently large. In other words, we can assign each product $\overrightarrow{v} \cdot \overrightarrow{c}$ a value independently of the others.

In SGNS model, we have
$$P((v,c)\in D)=\sigma(\overrightarrow{v} \cdot \overrightarrow{c})=\frac{1}{1+e^{-\overrightarrow{v} \cdot \overrightarrow{c}}}$$

Suppose we choose $k$ negative samples for each node-context pair $(v,c)$ according to the distribution $P_D(c_N)=\frac{\#(c_N)}{|D|}$. Then the objective function for SGNS can be written as
\begin{align*}
l&=\sum_{v\in V}\sum_{c\in V_C}\#(v,c)(\log \sigma(\overrightarrow{v} \cdot \overrightarrow{c})+k\cdot\mathbb{E}_{c_N\sim P_D}[\log \sigma(-\overrightarrow{v} \cdot \overrightarrow{c})])\\
&=\sum_{v\in V}\sum_{c\in V_C}\#(v,c)(\log \sigma(\overrightarrow{v} \cdot \overrightarrow{c})+k\cdot\sum_{v\in V}\#(v) \sum_{c_N\in V_C}\frac{\#(c_N)}{|D|}\log \sigma(-\overrightarrow{v} \cdot \overrightarrow{c})\\
&=\sum_{v\in V}\sum_{c\in V_C} (\#(v,c)(\log \sigma(\overrightarrow{v} \cdot \overrightarrow{c})+k\cdot\#(v)\cdot\frac{\#(c)}{|D|} \log \sigma(-\overrightarrow{v} \cdot \overrightarrow{c})
\end{align*}

Denote $x=\overrightarrow{v} \cdot \overrightarrow{c}$. By solving $\frac{\partial l}{\partial x}=0$, we have
$$\overrightarrow{v} \cdot \overrightarrow{c} = x = \log \frac{\#(v,c)\cdot |D|}{\#(v)\cdot \#(c)} -\log k$$

Thus we have $M_{ij}=\log \frac{\frac{\#(v_i,c_j)}{ |D|}}{\frac{\#(v_i)}{|D|}\cdot \frac{\#(c_j)}{|D|}} -\log k$. $M_{ij}$ can be interpreted as Pointwise Mutual Information(PMI) of node-context pair $(v_i,c_j)$ shifted by $\log k$.

\subsection{Softmax}
Since both Negative Sampling and Hierarchical Softmax are variants of softmax, we pay more attention to softmax model and give a further discussion in next section. We also assume that the values of $\overrightarrow{v} \cdot \overrightarrow{c}$ are independent.

In softmax model,
$$P((v,c)\in D)=\frac{e^{\overrightarrow{v} \cdot \overrightarrow{c}}}{\sum_{c'\in V_C} e^{\overrightarrow{v} \cdot \overrightarrow{c'}}}$$

And the objective function is
\begin{displaymath}
l=\sum_{v\in V}\sum_{c\in V_C} \#(v,c)\cdot \log \frac{e^{\overrightarrow{v} \cdot \overrightarrow{c}}}{\sum_{c'\in V_C} e^{\overrightarrow{v} \cdot \overrightarrow{c'}}}
\end{displaymath}

After extracting all terms associated to $\overrightarrow{v} \cdot \overrightarrow{c}$ as $l(v,c)$, we have
$$
l(v,c)=\#(v,c)\log \frac{e^{\overrightarrow{v} \cdot \overrightarrow{c}}}{\sum_{c'\in V_C, c'\neq c} e^{\overrightarrow{v} \cdot \overrightarrow{c'}}+e^{\overrightarrow{v} \cdot \overrightarrow{c}}}
+\sum_{\tilde{c}\in V_C, \tilde{c}\neq c}\#(v,\tilde{c}) \log\frac{e^{\overrightarrow{v} \cdot \overrightarrow{\tilde{c}}}}{\sum_{c'\in V_C, c'\neq c} e^{\overrightarrow{v} \cdot \overrightarrow{c'}}+e^{\overrightarrow{v} \cdot \overrightarrow{c}}}
$$
\newline
Note that $l=\frac{1}{|V_C|}\sum_{v\in V}\sum_{c\in V_C} l(v,c)$. Denote $x=\overrightarrow{v} \cdot \overrightarrow{c}$. By solving $\frac{\partial l}{\partial x}=0$ for all such x, we have
$$\overrightarrow{v} \cdot \overrightarrow{c} = x = \log \frac{\#(v,c)}{\#(v)} + b_v$$
where $b_v$ can be any real constant since it will be canceled when we compute $P((v,c)\in D)$. Thus we have $M_{ij}= \log \frac{\#(v_i,c_j)}{\#(v_i)} + b_{v_i}$. We will discuss what $M_{ij}$ represents in next section.
\section{Discussion}
It is clear that the method of sampling node-context pairs will affect matrix $M$. In this section, we will discuss $\frac{\#(v)}{|D|}$, $\frac{\#(c)}{|D|}$ and $\frac{\#(v,c)}{\#(v)}$ based on an ideal sampling method for DeepWalk algorithm.

Assume the graph is connected and undirected and window size is $t$. We can easily generalize this sampling method to directed graph by only adding $(RW_i,RW_j)$ into $D$.
\begin{algorithm}
\caption{Ideal node-context pair sampling algorithm}
\begin{algorithmic}
\STATE Generate an infinite long random walk $RW$.
\STATE Denote $RW_i$ as the node on position i of $RW$, where $i=0,1,2,\dots$
\FOR{$i=0,1,2,\dots$}
    \FOR{$j\in [i+1, i+t]$}
        \STATE add $(RW_i,RW_j)$ into $D$
        \STATE add $(RW_j,RW_i)$ into $D$
    \ENDFOR
\ENDFOR
\end{algorithmic}
\end{algorithm}

Each appearance of node $i$ will be recorded $2t$ times in $D$ for undirected graph and $t$ times for directed graph. Thus we can figure out that $\frac{\#(v_i)}{|D|}$ is the frequency of $v_i$ appears in the random walk, which is exactly the PageRank value of $v_i$. Also note that $\frac{\#(v_i,v_j)}{\#(v_i)/2t}$ is the expectation times that $v_j$ is observed in left/right $t$ neighbors of $v_i$. 

Denote the transition matrix in PageRank algorithm be A. More formally, let $d_i$ be the degree of node $i$. $A_{ij}=\frac{1}{d_i}$ if $(i,j)\in E$ and $A_{ij}=0$ otherwise. We use $e_i$ to denote a $|V|$-dimension row vector where all entries are zero except the $i$-th entry is 1.

 Suppose that we start a random walk from node $i$ and use $e_i$ to denote the initial state. Then $e_iA$ is the distribution over all the nodes where $j$-th entry is the probability that node $i$ walks to node $j$. Hence $j$-th entry of $e_iA^t$ is the probability that node $i$ walks to node $j$ at exactly $t$ steps. Thus $[e_i(A+A^2+\dots+A^t)]_j$ is the expectation times that $v_j$ appears in right $t$ neighbors of $v_i$.

 Hence

 $$\frac{\#(v_i,v_j)}{\#(v_i)/2t}=2[e_i(A+A^2+\dots+A^t)]_j$$
  $$\frac{\#(v_i,v_j)}{\#(v_i)}=\frac{[e_i(A+A^2+\dots+A^t)]_j}{t}$$

This equality also holds for directed graph.

By setting $b_{v_i}=\log 2t$ for all $i$, $M_{ij}=\log\frac{\#(v_i,v_j)}{\#(v_i)/2t}$ is logarithm of the expectation times that $v_j$ appears in left/right $t$ neighbors of $v_i$.

By setting $b_{v_i}=0$ for all $i$, $M_{ij}=\log\frac{\#(v_i,v_j)}{\#(v_i)}=\log \frac{[e_i(A+A^2+\dots+A^t)]_j}{t}$ is logarithm of the average probability that node $i$ randomly walks to node $j$ in $t$ steps.

\bibliographystyle{abbrv}
\bibliography{deepwalk}

\end{document}